\documentclass[10pt, a4paper]{article}

\usepackage[final]{lrec2026} 

\usepackage{tikz}
\usetikzlibrary{decorations.pathreplacing,calc,arrows.meta}

\usetikzlibrary{decorations.pathreplacing,calc,positioning}

\usepackage{graphicx} 
\usepackage[dvipsnames]{xcolor}
\usepackage[dvipsnames,x11names,svgnames]{xcolor}
\usepackage{amsmath}
\usepackage{float}

\title{Is Cross-Lingual Transfer in Bilingual Models Human-Like? A Study with Overlapping Word Forms in Dutch and English}

\name{Iza \v{S}krjanec$^{1,2}$, Irene Elisabeth Winther$^{3}$, Vera Demberg$^{1}$, Stefan L. Frank$^{3}$} 

\address{$^{1}$Saarland University, Germany $^{2}$Zuse School ELIZA, Germany $^{3}$Radboud University, the Netherlands \\
         \{skrjanec, vera\}@lst.uni-saarland.de, \{irene.winther, stefan.frank\}@ru.nl\\
         }

\abstract{
Bilingual speakers show cross-lingual activation during reading, especially for words with shared surface form. Cognates (friends) typically lead to facilitation, whereas interlingual homographs (false friends) cause interference or no effect. We examine whether cross-lingual activation in bilingual language models mirrors these patterns. We train Dutch-English causal Transformers under four vocabulary-sharing conditions that manipulate whether (false) friends receive shared or language-specific embeddings. Using psycholinguistic stimuli from bilingual reading studies, we evaluate the models through surprisal and embedding similarity analyses. The models largely maintain language separation, and cross-lingual effects arise primarily when embeddings are shared. In these cases, both friends and false friends show facilitation relative to controls. Regression analyses reveal that these effects are mainly driven by frequency rather than consistency in form-meaning mapping. Only when just friends  share embeddings are the qualitative patterns of bilinguals reproduced. Overall, bilingual language models capture some cross-linguistic activation effects. However, their alignment with human processing seems to critically depend on how lexical overlap is encoded, possibly limiting their explanatory adequacy as models of bilingual reading. 
 \\ \newline  }

\begin{document}

\maketitleabstract

\section{Introduction}
Compared to speakers of a single language, bilingual speakers process some linguistic phenomena differently during reading, a notable example being words with the same surface form in multiple languages. For cognate words, the surface form as well as the meaning is equivalent or very similar across languages (e.g.~\textit{winter} in Dutch and English). For interlingual homographs (henceforth, false friends) the surface form is the same, but the meaning is not (e.g.~\textit{brand} means `fire' in Dutch).

Empirical studies of human reading find that bilinguals process cognate words faster than non-cognate control words \citep{Libben2009BilingualLA,Bultena2014CognateEI,LAURO2017217}. This cognate facilitation effect has not been observed in monolinguals, suggesting that it arises specifically from the coexistence of two languages within a single speaker.

For reading of false friends, studies report either slower reading of false friends than control words or no difference in bilingual speakers \citep{Libben2009BilingualLA, Titone2011BilingualLA, Pivneva2014, Hoversten2015ATC}. Slower reading of false friends in bilinguals is thought to reflect interference caused by diverging form-meaning mappings across languages. 

\begin{figure}[!t]
  \centering
  \begin{tikzpicture}[
    scale=0.75, transform shape, 
    every node/.style={font=\footnotesize}
  ]

  \def\Rorange{1.55}
  \def\Rblue{1.35}

  \tikzset{
    dutchlabel/.style={font=\footnotesize\bfseries, text=orange!80!black},
    englishlabel/.style={font=\footnotesize\bfseries, text=blue!80!black},
    condlabel/.style={font=\footnotesize\bfseries\normalfont, align=center}
  }

  \def\xsep{5.2}
  \def\ysep{5.2}

  \def\condy{-2.0}

  \begin{scope}[shift={(0,0)}]
    \def\xorange{-0.60}
    \def\xblue{0.60}

    \draw[orange, thick, fill=orange!40, fill opacity=0.35] (\xorange,0) circle (\Rorange);
    \draw[blue,   thick, fill=blue!40,   fill opacity=0.35] (\xblue,0)   circle (\Rblue);

    \node[dutchlabel]   at (\xorange-0.8,1.77) {Dutch};
    \node[englishlabel] at (\xblue+0.4,1.77)   {English};

    \node[align=center] at (0,0) {is\\ winter\\ brand\\ want\\ *};
    \node at (-1.4,-0.00) {kogel};
    \node at (1.43,-0.00)  {prison};

    \node[condlabel] at (0,\condy) {(A) \textbf{\textcolor{ForestGreen}{Full Overlap}}};
  \end{scope}

  \begin{scope}[shift={(\xsep,0)}]
    \def\xorange{-0.75}
    \def\xblue{0.75}

    \draw[orange, thick, fill=orange!40, fill opacity=0.35] (\xorange,0) circle (\Rorange);
    \draw[blue,   thick, fill=blue!40,   fill opacity=0.35] (\xblue,0)   circle (\Rblue);

    \node[dutchlabel]   at (\xorange-0.8,1.77) {Dutch};
    \node[englishlabel] at (\xblue+0.4,1.77)   {English};

    \node[align=center] at (0.14,0) {winter\\ is\\ *};

    \node at (-1.45,-0.00) {kogel};
    \node at (-1.2,0.70) {brand$_{NL}$};
    \node at (-1.1,-0.80) {want$_{NL}$};
    \node at (1.35,0.65) {want$_{EN}$};
    \node at (1.5,0.05) {brand$_{EN}$};
    \node at (1.2,-0.65) {prison};

    \node[condlabel] at (0,\condy) {(B) \textbf{\textcolor{Teal}{Friends Overlap}}};
  \end{scope}

  \begin{scope}[shift={(0,-\ysep)}]
    \def\xorange{-0.75}
    \def\xblue{0.75}

    \draw[orange, thick, fill=orange!40, fill opacity=0.35] (\xorange,0) circle (\Rorange);
    \draw[blue,   thick, fill=blue!40,   fill opacity=0.35] (\xblue,0)   circle (\Rblue);

    \node[dutchlabel]   at (\xorange-0.8,1.77) {Dutch};
    \node[englishlabel] at (\xblue+0.4,1.77)   {English};

    \node[align=center] at (0.14,-0.1) {brand\\ want\\ *};

    \node at (-1.45,-0.00) {kogel};
    \node at (-1.2,0.70) {winter$_{NL}$};
    \node at (-1.1,-0.80) {is$_{NL}$};
    
    \node at (1.35,0.65) {winter$_{EN}$};
    \node at (1.5,0.05) {is$_{EN}$};
    \node at (1.2,-0.65) {prison};

    \node[condlabel] at (0,\condy) {(C)\ \textbf{\textcolor{Gold3}{False Friends Overlap}}};
  \end{scope}

  \begin{scope}[shift={(\xsep,-\ysep)}]
    \def\xorange{-0.95}
    \def\xblue{0.95}

    \draw[orange, thick, fill=orange!40, fill opacity=0.35] (\xorange,0) circle (\Rorange);
    \draw[blue,   thick, fill=blue!40,   fill opacity=0.35] (\xblue,0)   circle (\Rblue);

    \node[dutchlabel]   at (\xorange-0.8,1.77) {Dutch};
    \node[englishlabel] at (\xblue+0.4,1.77)   {English};

    \node[align=center] at (0.15,0) {*};

    \node at (-1.6,0.3) {kogel};
    \node at (-1.2,0.80) {winter$_{NL}$};
    \node at (-1.7,-0.19) {is$_{NL}$};
    \node at (-1.1,-0.70) {brand$_{NL}$};
    \node at (-0.8,-1.20) {want$_{NL}$};


    \node at (1.2,0.9) {winter$_{EN}$};
    \node at (1.35,0.4) {is$_{EN}$};
    \node at (1.5,0.05) {prison};
    \node at (1.5,-0.35) {want$_{EN}$};
    \node at (1.2,-0.8) {brand$_{EN}$};

    \node[condlabel] at (0,\condy) {(D) \textbf{\textcolor{Crimson}{Minimal Overlap}}};
  \end{scope}

  \end{tikzpicture}

  \caption{Vocabulary conditions. \textbf{\textcolor{ForestGreen}{Full overlap (A)}}: All identical surface forms that appear in Dutch and English are shared between the two languages, each represented by a single embedding. \textbf{\textcolor{Teal}{Friends Overlap (B)}}: Only cognates and loan words are shared. \textbf{\textcolor{Gold3}{False Friends Overlap (C)}}: Only false friends are shared. \textbf{\textcolor{Crimson}{Minimal Overlap (D)}}: Only punctuation and named entities are shared, while other tokens have language-specific embeddings.  Across all conditions, punctuation and named entities are shared (denoted as $*$).}
  \label{fig:venn-conditions}
\end{figure}
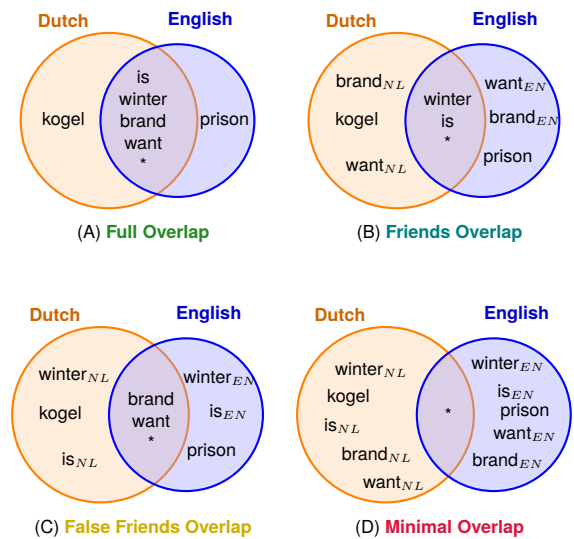

Both facilitation and interference effects have been attributed to cross-language activation \citep[e.g.][]{Dijkstra2002TheAO,KROLL2012229}. 
Explaining these cross-lingual interactions requires explicit and testable computational models of how the two languages interact within a single cognitive system \citep{Frank2021TowardCM}.
Neural language models (LMs) are strong candidates as computational models of second language learning and bilingual reading \citep{yadavalli-etal-2023-slabert, oba-etal-2023-second, aoyama-schneider-2024-modeling, constantinescu-etal-2025-investigating}.

In Natural Language Processing, cross-lingual transfer is often desired when structural or lexical similarities or simply lexical overlap between languages can be leveraged to improve downstream performance \citep[e.g.,][]{kallini-etal-2025-false, K2019CrossLingualAO, wu-dredze-2019-beto, pires-etal-2019-multilingual}. In contrast, this work evaluates bilingual LMs for cross-lingual transfer between Dutch and English and explores under what conditions this transfer seems human-like (if at all). We test LMs for lexical processing via vocabulary manipulation in four conditions (Figure \ref{fig:venn-conditions}), controlling how overlapping word forms are treated in the vocabulary. In the \textcolor{Teal}{Friends Overlap} condition, each word that is a \textit{friend} (a cognate or a loan word) is assigned a single, language-unspecific embedding allowing both Dutch and English to inform it. In the \textcolor{Gold3}{False Friends Overlap} setting, the same holds for words that are false friends (but not friends). The other two conditions enforce either sharing of each form-overlapping word (\textcolor{ForestGreen}{Full Overlap}) or a nearly complete separation of  languages in the embedding space (\textcolor{Crimson}{Minimal Overlap}).

We evaluate each vocabulary condition through the lens of two signals important in LM training as well as cross-lingual activation in humans: word frequency and language context (i.e.~the language of the sentence). 

Our results show that bilingual models generally separate the two languages, except when they are explicitly tied by a shared embedding of a word. In that case, the models show cross-lingual effects, specifically facilitation effects for friends as well as false friends compared to their respective control words. We find that this facilitation is driven by word frequency, and is not influenced by form-meaning mapping across languages. The only condition that can explain human data is \textcolor{Teal}{Friends Overlap} (i.e.~facilitation for friends, but not for false friends). This suggests that while LMs can reproduce certain cross-lingual activation patterns, their behavior aligns with human bilingual reading only under specific vocabulary conditions\footnote{Our code is made available at \url{https://github.com/izaskr/cross_lingual_transfer_dutch_english_forms}.}.


\section{Background and Related Work}

\subsection{Vocabulary Design} \label{sec:rw_vocab}
A large number of related studies use an equivalent of the \textcolor{ForestGreen}{Full Overlap} condition when training a bilingual LM, so each overlapping surface form is shared between languages \citep{winther-etal:2021-cumulative, Roslund2022ModelingSP, oba-etal-2023-second, constantinescu-etal-2025-investigating}. Our vocabulary conditions are inspired by \citet{kallini-etal-2025-false}, who experiment with vocabulary manipulations to find that, for downstream tasks such as natural language inference and question answering, any sharing is beneficial (even that of false friends) in contrast to no overlap. \citet{aoyama-schneider-2024-modeling} reset the embedding layer before starting L2 acquisition, which essentially does not allow cross-lingual transfer between overlapping word forms. 

In terms of the share of first (L1) and second (L2) language, some studies used a balanced proportion where each language has the same budget of training tokens \citep{oba-etal-2023-second, kallini-etal-2025-false,constantinescu-etal-2025-investigating}, while others simulated larger L1 exposure compared to L2 \citep{Roslund2022ModelingSP, constantinescu-etal-2025-investigating}. \citet{winther-etal:2021-cumulative} observe that the cognate facilitation effect in LMs trained on Dutch-English or Norwegian-English depends on the portion of L1/L2 and the presentation order of the languages, whereby an LM that is first trained on the L1 (75\% training samples), and then on the L2 at the end of each epoch exhibits lower surprisal for cognates relative to non-cognates, mirroring the facilitation effect observed in bilinguals.

\subsection{Role of Frequency and Context}
Language models learn more frequent words earlier and better \citep{chang-bergen-2022-word, razeghi-etal-2022-impact}. This provides a relevant link to cross-lingual transfer in bilinguals.
According to the cumulative frequency hypothesis \citep{Voga2007, strijkers-costa-thierry-tracking, midgley-effects-cognate-erp}, the cognate facilitation effect in bilinguals stems from exposure to cognates regardless of the language: exposure strengthens the word's lexical representation due to the same form-meaning mapping. It is unclear how this fares with false friends, where the form-meaning mapping diverges across languages, i.e. whether we can expect cross-lingual transfer to be influenced by frequency.

Aside from word frequency, context is an important cue in LM training and inference. LMs learn how to use context information to disambiguate lexically ambiguous words within one language \citep{pilehvar-camacho-collados-2019-wic, raganato-etal-2020-xl}. One question that arises is whether a bilingual LM can use the language of the preceding context as a cue to process an overlapping word form without influence/transfer from the other language, or whether the fact that an overlapping form occurs in two language contexts makes it more ambiguous and less predictable.

\section{Manipulation of Vocabulary Sharing}

We design 4 conditions for the LM vocabulary to test the role of sharing/separation of embeddings of words with the same surface form across Dutch and English. When a word form is \textit{shared} (see intersections in Figure \ref{fig:venn-conditions}), it has a single word embedding. If this form happens to appear in both Dutch and English, then samples from both languages contribute to the embedding during training. In case of \textit{separation} (see the areas outside of the intersection in Figure \ref{fig:venn-conditions}), each word form is encoded for the language of the sentence it is in. Their embeddings are informed only by the samples in the respective language.

In condition \textcolor{ForestGreen}{Full overlap (A)}, there is no active intervention: each word form that appears in both languages has a single, language-unspecific embedding. In other conditions, we manipulate what is shared and what is separated. 

In condition \textcolor{Teal}{Friends Overlap (B)}, only friends (cognate and loan words) across the two languages are in the intersection. This condition is motivated by hypotheses that cognate words might share orthographic and (partially) semantic representations in a bilingual's mental lexicon, while false friends are represented by two different orthographic representations \citep{Dijkstra2002TheAO,LemhoeferDijkstra2004}. All other words have language-specific embeddings. 

Only false friends are placed in the intersection in condition \textcolor{Gold3}{False Friends Overlap (C)}. Each false friend has a single entry in the vocabulary despite different meanings across languages (e.g.,~ \textit{brand} means `fire' in Dutch). 

Finally, in \textcolor{Crimson}{Minimal Overlap (D)}, the word forms from the two languages are completely separated given the sentence language. A language-specific processing is assumed, regardless of overlapping form/meaning across languages.

Note that punctuation and named entities are placed in the intersection across all conditions. They are often not language-specific, refer to the same extra-linguistic entity and are thus not a part of our manipulation. Named entities were identified by spaCy's xx\_ent\_wiki\_sm model\footnote{\url{https://spacy.io/models/xx}} prior to tokenizer and LM training. 

\begin{figure}
    \centering
    \includegraphics[page=1,width=0.5\textwidth]{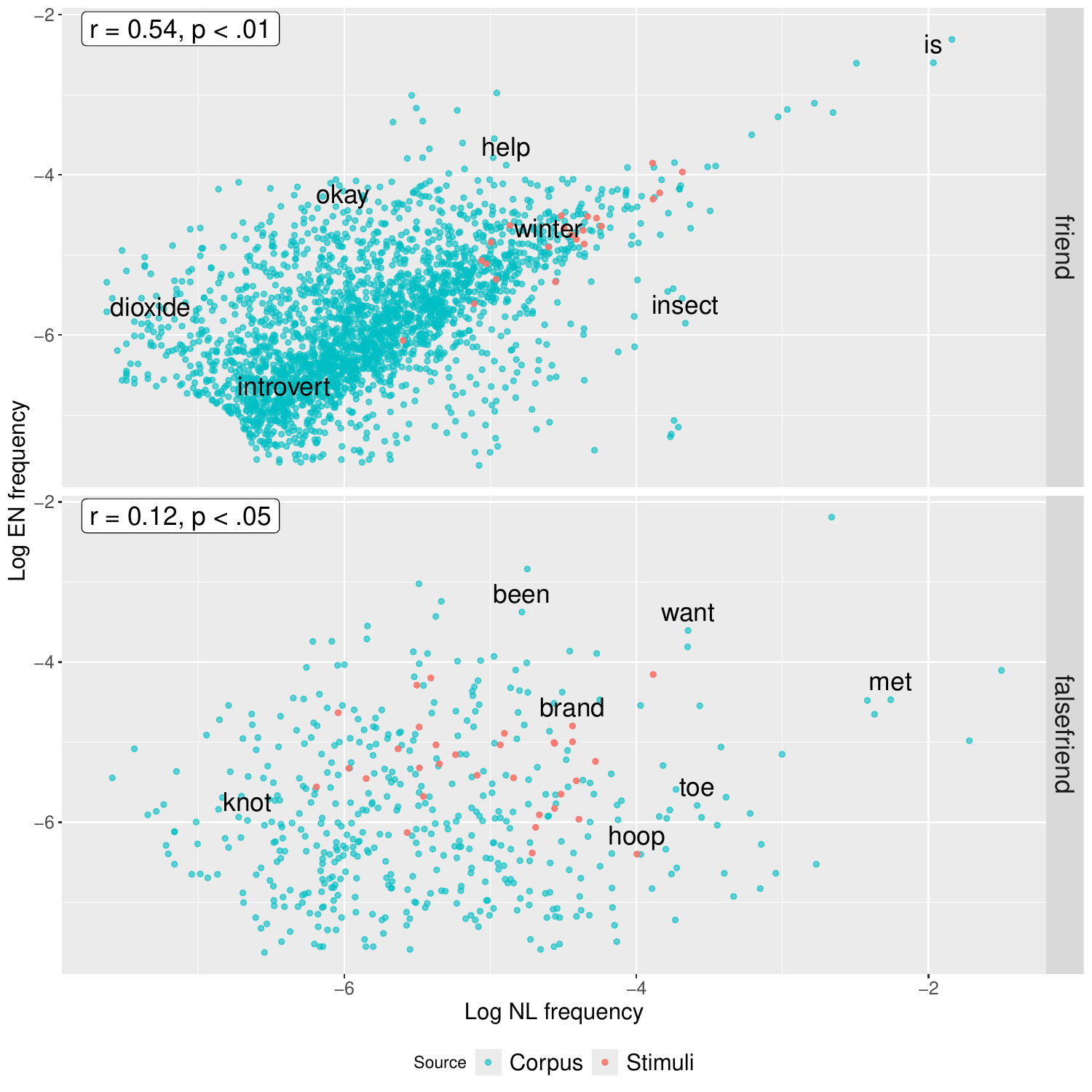}
    \caption{The frequency of each word in the Dutch and English portion of the LM training data. The frequencies are generally positively correlated: a word form that is frequent in Dutch is frequent in English, but more so for \textit{friends} than \textit{false friends}. The plot includes (false) friend words from psycholinguistic stimuli \citep{Bultena2014CognateEI, huisman-ma-thesis-ih} as well as other compiled lists and our training data (here labeled as Corpus).} 
    \label{fig:freq_train_data}
\end{figure}


\section{Implementation}

\subsection{Training Data}
To imitate human language exposure, we train our LMs on a corpus of diverse genres: 49\% of tokens in the data come from non-fiction text (Wikipedia\footnote{\url{https://dumps.wikimedia.org/nlwiki/latest}}), 26\% from transcribed scripted speech (OpenSubtitles\footnote{\url{https://opus.nlpl.eu/OpenSubtitles/nl&en/v2024/OpenSubtitles}} and TedTalks\footnote{\url{https://object.pouta.csc.fi/OPUS-NeuLab-TedTalks/v1/tmx/en-nl.tmx.gz}}), and 25\% from web-crawled data \citep[CC100, ][]{conneau-etal-2020-unsupervised}\footnote{\url{https://data.statmt.org/cc-100}}. In total, each training corpus had about 400 million tokens. To simulate an unbalanced late Dutch-English bilingual, the training data consists mostly of Dutch (75\% tokens), while the rest is English. In each epoch, Dutch samples were presented first, followed by English ones.

\subsection{Overlapping Word Forms} \label{sec:overlapping_words}
To obtain a broad coverage of words sharing only form, or both form and meaning between Dutch and English, we annotated a list of words that appear in both the Dutch and English corpora as friends (cognates or loan words) and false friends (interlingual homographs). We joined our list with existing ones \citep{Bultena2014CognateEI,Poort-2019, lefever-etal-2020-identifying, huisman-ma-thesis-ih} to obtain a set of 2,806 friends and 511 false friends, both referring to word types without named entities. The Dutch and English frequencies of each word have a significant positive correlation (Figure \ref{fig:freq_train_data}). 

Some words are homographs within one language and can be both a friend and a false friend (e.g.~\textit{monster} in Dutch can have the English meaning as well as `sample'). These words are excluded from our vocabulary manipulation.

Class annotations are available for a subset of the items, comprising of 1,598 friends and 379 false friends: all cognate and loan words are annotated as content words in both languages. Among false friends, 4\% have different classes in Dutch and English (e.g.~\textit{toe} and \textit{met} are content words in English, but function words in Dutch). 



\subsection{Psycholinguistic Stimuli} \label{sec:exp_stimuli}
We evaluate the LMs on stimuli from two reading studies, in which the participants were late bilinguals with Dutch as their first language and English as their second. In the \textit{friend study} \citep{Bultena2014CognateEI}, participants read English sentences with target words that were either Dutch-English friends (cognates) (1a) or non-cognate controls (1b)\footnote{We only use items where the cognate is a form-identical noun.}.

\begin{itemize}
    \item[1a.] The residents dislike the \textit{winter} for the ...
    \item[1b.] The residents dislike the \textit{prison} for the ...
\end{itemize}

The friend and control targets were matched with respect to word length and English word frequency. In total, the stimuli consisted of 22 item sentences (with two conditions per item). The context preceding the target words was designed to not be semantically constraining and biased towards the meanings of target words. The analysis of reading times by Dutch-English bilinguals revealed a cognate facilitation effect, with faster reading for cognates than the corresponding control words.

The stimuli in the \textit{false-friend study} \citep{huisman-ma-thesis-ih} include word pairs of false friends and controls that were presented to participants in Dutch sentences, that is, in their first language, in a self-paced reading paradigm. Again, each sentence had two conditions: with a false friend (2a) or with a language-unique control word (2b):

\begin{itemize}
    \item[2a.] De beelden van de \textit{brand} zullen hen ...
    \item[2b.] De beelden van de \textit{kogel} zullen hen ...
\end{itemize}

All target words were nouns and all false friends were form-identical between Dutch and English. Control words were paired with false friends to match them on word length and Dutch and English word frequency. The stimuli by \citet{huisman-ma-thesis-ih} consists of 32 item sentences, all designed to be semantically non-constraining in the context preceding the target words. While this study did not find significant reading time differences between false friends and control words in bilinguals (and thus no cross-language activation), the manipulation provides a valuable evaluation of the LMs tested here, particularly due to the lack of prior work on false-friend reading in Dutch-English bilinguals.

\subsection{Tokenizers}
The tokenization process included multiple components. All words annotated as \textit{friends} or \textit{falsefriends} (Section \ref{sec:overlapping_words}) as well as the language-unique control words (Section \ref{sec:exp_stimuli}) were tokenized as single subwords across all conditions to avoid biases from multi-subword segmentation \citep{lesci-etal-2025-causal}.
We train a byte-level BPE tokenizer on combined Dutch-English train set (64k vocabulary, minimum frequency of 2) and used it to tokenize all remaining words. We also train a byte-level BPE tokenizer on named entities (10k vocabulary, minimum frequency of 2).

\subsection{Transformer Language Models}
For each vocabulary manipulation, we train a separate Transformer model with the causal language modeling task, following the GPT2-small configuration. In a single epoch, the model was first presented Dutch samples (300 million tokens), followed by English ones (100 million tokens) creating a sequential regime. Each LM was trained for 2 epochs. See Appendix \ref{app:vocab_lm_sizes} for vocabulary sizes and parameter counts under different conditions. Appendix \ref{app:loss_plots} shows loss curves on the training and test sets.


\section{Vocabulary Interventions in Bilingual LMs}


\subsection{Language Context} 
We first explore how (dis)similar the preceding contexts of the same surface form are across Dutch and English. If the LM ``understands'' that a friend word carries the same meaning in Dutch and English, this should be reflected in the similarity of its contextual and word embeddings across languages. In the same vein, if the LM represents the language-specific meanings of false friends, their context and word embeddings should be dissimilar as well. In conditions that assign each overlapping word form a shared embedding (\textcolor{Teal}{Friends Overlap}  for friends; \textcolor{Gold3}{False Friends Overlap} for false friends; \textcolor{ForestGreen}{Full Overlap} for both), it might be easier for the LM to learn friends as cross-lingual meaning equivalents, but it might be more difficult to tease apart the language-specific meanings of false friends.


\paragraph{Method.} For each word $t$ in the stimuli set of friends and false friends $T$, we sample a set of 500 sentences containing that word from the training data from each language (yielding 1,000 sentences per word in total). For each sentence in the sample, we calculate the mean-pooled embedding of the preceding context of word $t$ (without the word itself).  For each language-specific subcorpus,  we average across these embeddings, obtaining $\mu_{C}^{NL}(t)$ and $\mu_{C}^{EN}(t)$.

For each sentence in the sample, we also obtain the contextualized embedding of the target word $t$ itself. We then average across the sentences per language, yielding $\mu_{W}^{NL}(t)$ and $\mu_{W}^{EN}(t)$ for Dutch and English sentences, respectively. 

We measure the similarity of these embeddings between languages using cosine similarity\footnote{Due to known issues with anisotropy in contextualized representations \citep{ethayarajh-2019-contextual}, we standardize the embeddings following \citet{timkey-van-schijndel-2021-bark}. We estimate a mean and standard deviation embedding from a sample of 30k instances and standardize the embeddings before pooling.}. The similarity between $\mu_{C}^{NL}(t)$ and $\mu_{C}^{EN}(t)$ tells us how similar the preceding contexts of the same word are across languages. In contrast, cosine similarity between $\mu_{W}^{NL}(t)$ and $\mu_{W}^{EN}(t)$ estimates the similarity of the target word embedding across languages. See Appendix \ref{app:diagrams_vectors} for illustrative diagrams.
We use the embeddings from layer 12. Results from layer 5 show very similar patterns.

\paragraph{Results. }

\begin{figure}
    \centering
    \includegraphics[page=1,width=0.5\textwidth]{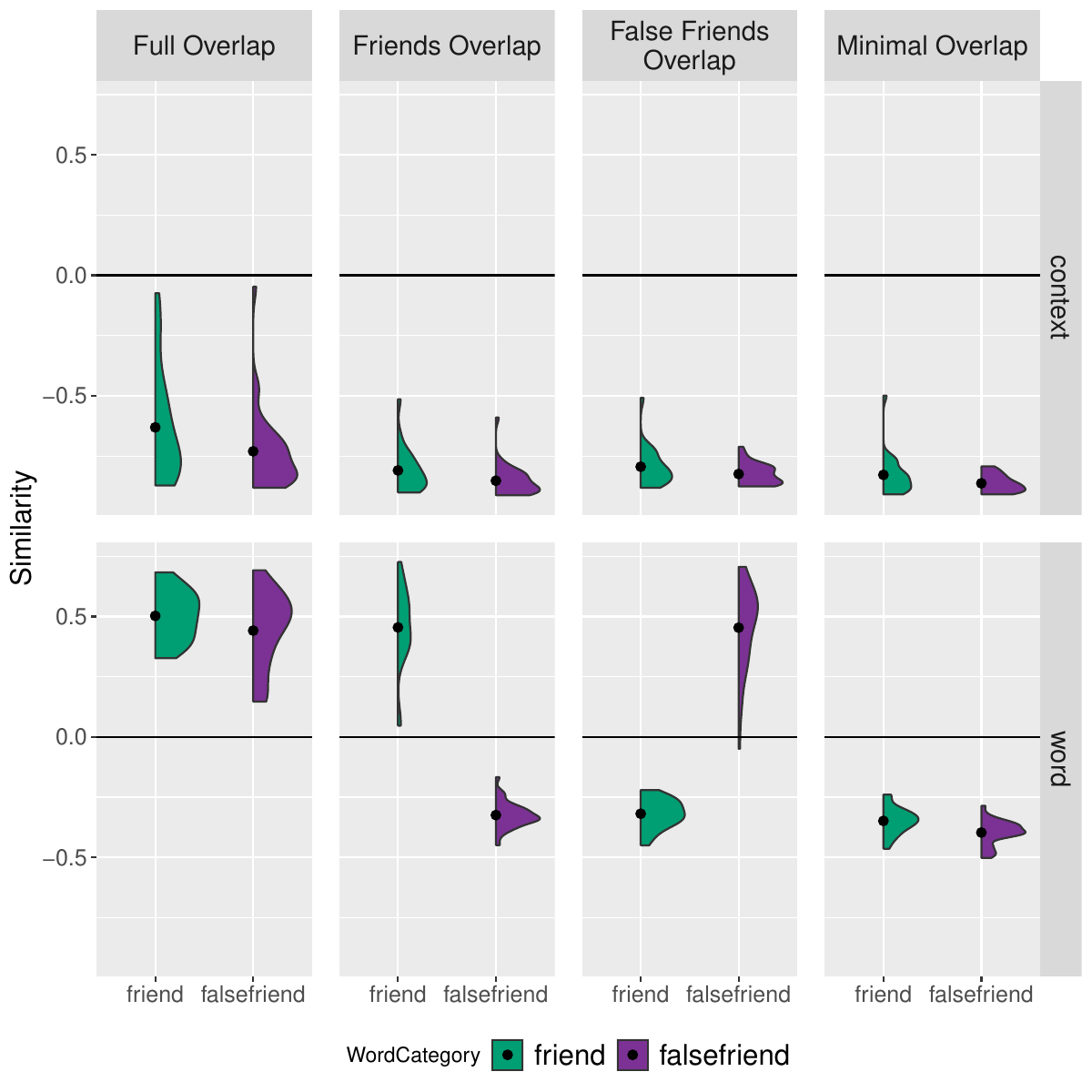}
    \caption{Cosine similarity between Dutch and English across the context and word-only embeddings for training data samples for target words (friends or false friends). A larger similarity means the LM considers Dutch and English embeddings more similar.} 
    \label{fig:contextual_sim}
\end{figure}

Figure \ref{fig:contextual_sim} shows that the contexts for the same word in Dutch and English are generally estimated as dissimilar and vary little across vocabulary conditions, i.e.~manipulating what is shared in the embedding space does not lead to more similarities or differences for the context preceding target words (upper panel). Focusing on the embeddings of the target words themselves (lower panel), the results indicate that separating the languages for overlapping word forms leads to lower similarity for the word affected by separation: in the \textcolor{Teal}{Friends Overlap} condition, false friends have language-specific vocabulary entries. The distance between Dutch and English sentences containing these words is larger than when each false friend has a single embedding. Friends, likewise, show a lower similarity when their embeddings are language-specific (\textcolor{Gold3}{False Friends Overlap} condition). In the \textcolor{Crimson}{Minimal Overlap} condition, the contexts as well as word embeddings stay apart and dissimilar across Dutch and English. Generally, this result shows that the contexts are coded as dissimilar between Dutch and English, and the target word embeddings are similar when each is represented by a single, language-unspecific embedding, no matter the form-meaning match in Dutch and English.


\subsection{Processing Effort via Surprisal} \label{sec:surprisal}

We further focus on the predictability of overlapping word forms in comparison to language-unique words given the same preceding sentence context by estimating their surprisal. We use word surprisal, $surp(w) = -\log_{2}p(w|context)$, as a computational correlate of processing effort as observed in humans with lower surprisal corresponding to easier processing \citep[e.g.][]{hale-2001-probabilistic, Levy2008ExpectationbasedSC, shain-large-scale}. 

Bilingual speakers tend to show different behavior for language-unique words and words with overlapping surface form. Specifically, reading a word with form overlap across languages might show patterns of cross-language activation: facilitation for friends (faster reading in comparison to controls), and interference for false friends (slower reading in comparison to controls). We expect lower surprisal for friends than controls, reflecting the ease of processing of friends over controls. In contrast, we expect lower surprisal for controls than false friends.

With respect to the vocabulary configurations, we examine how sharing/separation of embeddings affects the form-meaning correspondence of overlapping word forms. On the one hand, in conditions where each overlapping surface forms has a single embedding, this word might be more ambiguous as it appears across Dutch and English contexts, possibly  reducing its predictability in a given sentence. If so, shared forms should have higher surprisal than language-unique controls. On the other hand, a shared embedding receives more training signal than two single ones and more than a frequency-matched control, which could result in reduced surprisal compared to controls. We use $\alpha=.05$ in all statistical tests.


\paragraph{Method.}

Using sentence stimuli from the \textit{friend study} and the \textit{false-friend} study (see Section \ref{sec:exp_stimuli}), we estimate surprisal for each target word. Importantly, in each item, the target word (either overlapping or control) is preceded by the same context. We test for effects of processing effort by comparing the surprisal of overlapping words to that of controls.

We use mixed-effects regression in \texttt{lme4} \citep{lme4} with surprisal as the outcome, and word category and word position in sentence (index) as fixed effects, and a random intercept for sentence pair (item)\footnote{Model structure: $surp \sim WordCategory + WordIndex + (1|Item)$. WordCategory is sum-coded (1 friend/falsefriend, $-$1 control).}. We fit a regression model for each vocabulary condition separately for the two studies, resulting in 8 regression models in total.

\paragraph{Results.}

\begin{figure*}[!h]
    \centering
    \includegraphics[width=\textwidth]{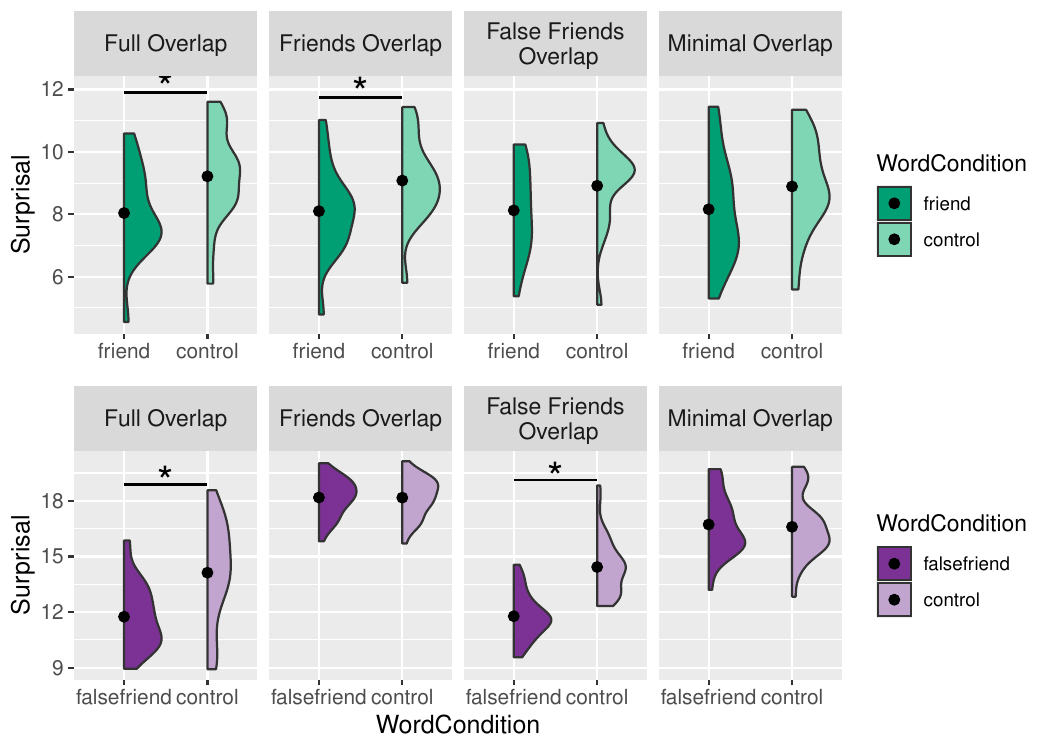}
    \caption{Surprisal estimates for friends/controls and false friends/controls from LMs trained under different vocabulary conditions. The symbol $*$ indicates a statistically significant difference ($p<.05$) between the two word types.}
    \label{fig:surprisal_both}
\end{figure*}

Figure \ref{fig:surprisal_both} shows surprisal estimates across the two word categories and vocabulary conditions. For \textit{friends}, we find that they have significantly lower surprisal than control words in two conditions: \textcolor{ForestGreen}{Full Overlap}, and \textcolor{Teal}{Friends Overlap}. In both conditions, each friend surface form has a single embedding, which increases the word's predictability relative to controls. For \textit{false friends}, we find that in the \textcolor{ForestGreen}{Full Overlap} and \textcolor{Gold3}{False Friends Overlap} conditions, their surprisal is significantly lower than that of controls. Again, in these conditions, the false friend's surface form has a single embedding.

Taken together, both types of overlapping word forms (friends and false friends) show smaller surprisal relative to language-unique controls when their surface forms are represented by a single shared embedding. These results indicate that embedding sharing increases the predictability of overlapping surface forms regardless of their form-meaning correspondence. Surprisingly, neither word category show lower predictability, either because of language context ambiguity or form-meaning divergence. 

\subsection{Frequency Benefit across Languages}

In bilingual training, overlapping surface forms receive additional exposure relative to language-unique controls due to their occurrence in both languages. This raises the question whether increased frequency drives improved predictability of words with overlapping surface forms across languages (i.e.~as per the cumulative frequency hypothesis) or whether their predictability might depend on the language of the sentence. This analysis builds on the facilitation effects observed in Section \ref{sec:surprisal} focusing on the effect of a word's frequency in Dutch vs. English on its surprisal. 

The two frequency measures may play different roles. In the \textit{friend study}, the target words appear in English sentences, making Dutch the ``other'' language. In the \textit{false-friend study}, the sentences are in Dutch, so English serves as the ``other'' language. This analysis examines whether both the sentence-language frequency ($\text{freq}_\text{S}(w)$) and the other-language frequency ($\text{freq}_\text{O}(w)$) influence surprisal of word $w$.

\paragraph{Method.}
Both friends and false friends exhibited facilitation relative to controls in vocabulary conditions where their embeddings were shared between Dutch and English. Thus, to compare the two frequency measures ($\text{freq}_\text{S}(w)$ and $\text{freq}_\text{O}(w)$), we consider the \textcolor{ForestGreen}{Full Overlap} and \textcolor{Teal}{Friends Overlap} conditions for the \textit{friend study}, and \textcolor{ForestGreen}{Full Overlap} and \textcolor{Gold3}{False Friends Overlap} for the \textit{false-friend study}. The frequency estimates of the target words are based on the Dutch and English portions of the LM training data; they are log-transformed and normalized. For each study, we start by fitting a regression model with surprisal estimated by word category, normalized word index and vocabulary condition\footnote{The maximal model that would converge was $surp \sim WordCategory + Vocabulary + WordIndex + (1|Item)$.}. In the next step, we add $\text{freq}_\text{S}$ to the model\footnote{$surp \sim freq_{S} + WordCategory + Vocabulary + WordIndex + (1|Item)$}, and finally $\text{freq}_\text{O}$\footnote{$surp \sim freq_{O} + freq_{S} + WordCategory + Vocabulary + WordIndex + (1|Item)$. Interaction terms did not improve the model fit for any regression models.}. We compare these regression models using likelihood ratio tests to determine whether the addition of the frequency terms provides a better fit.

\paragraph{Results.}
The results for the \textit{friend study} show a main effect of word category as facilitation for friends over controls ($\beta = -.5$, $p<.01$). Upon adding word frequency in English ($\text{freq}_\text{S}$), the word category effect was still negative, but not a reliable effect ($\beta = -.2$, $p=.06$). There is a main effect of English frequency ($\beta = -1.1$, $p<.01$), indicating that higher frequency leads to smaller surprisal. Finally, including Dutch word frequency ($\text{freq}_\text{O}$), we observe that English frequency is still a significant predictor ($\beta = -1.2$, $p<.01$), but Dutch frequency is not ($p=.9$) and neither is word category ($p=.4$). The model comparison revealed that including Dutch word frequency in the model does not contribute to the fit ($\chi^2 (1)= .002, p = .97$).

For the \textit{false-friend study}, we find a facilitation effect for false friends compared to controls ($\beta=-1.3$, $p<.01$). When Dutch frequency ($\text{freq}_\text{S}$) is included in the model, there is still a significant facilitation effect for false friends ($\beta=-1.2$, $p<.01$) as well as a significant main effect of Dutch frequency ($\beta=-.5$, $p<.01$). Finally, when English frequency ($\text{freq}_\text{O}$) is added to the model, Dutch frequency remains a significant predictor ($\beta=-.5$, $p<.05$), and English frequency is significant as well ($\beta=-.7$, $p<.05$), but the difference between false friends and controls is no longer significant ($\beta=-.6$, $p=.06$). Model comparison further indicates that including English word frequency significantly improves model fit ($\chi^2(1) = 4.49, p = .034$). 

Overall, these results indicate that effects of bilingual exposure differ between L1 and L2. In the \textit{friend study}, the facilitation effect for friends over controls is accounted for by sentence-language frequency (English), with seemingly no contribution from the other-language frequency (Dutch). However, correlation of fixed effects sheds light on this: there is a large correlation between word category and Dutch frequency ($r=-.9$), essentially signaling to the regression model that words that do not appear in Dutch (i.e.~have a Dutch frequency of 0) are English control words. For friends, the English and Dutch frequencies are highly correlated in the model ($r=-.7$), as well as in the data (see Figure \ref{fig:freq_train_data}). This means that English frequency sufficiently predicts variance between control and friend words. The variance that could be explained by Dutch frequency is already explained by the English one, making Dutch frequency redundant. 

In contrast, the facilitation effect found for the \textit{false-friend study} seems to be driven by both the frequency in the language of the sentence and frequency in the ``other'' language. In this model, word category and English frequency are correlated as well ($r=-.9$), again pertaining to Dutch control words that have a 0 frequency in English. However, the model correlation between Dutch and English false friends frequency is weak ($r=-.15$, see also Figure \ref{fig:freq_train_data} for data correlation). This is not surprising since in the case of false friends, the same surface form is mapped to different meanings and possibly to different parts of speech, with different distributional patterns across Dutch and English. We hypothesize that this is the reason both Dutch and English frequencies remain significant in the \textit{false-friend study}.

Generally, our results provide support for the cumulative frequency hypothesis for both friend and false friend words. Both languages contribute to the facilitation effect, but for friends it is harder to disentangle the effects of Dutch and English as frequency sources due to their high correlation, especially across highly similar languages \citep{schepens_cross-language-distributions}.

\section{Discussion }

This work asked whether cross-lingual transfer in bilingual Transformer LMs produces patterns that resemble cross-lingual activation in bilingual readers, focusing on words with overlapping surface forms across Dutch and English. To test this, we trained LMs on L1 Dutch and L2 English, and evaluated the models on experimental stimuli from studies of Dutch-English bilinguals, which included words with overlapping surface forms between the two languages. Designing a cognitively plausible vocabulary in bilingual LMs is an open question. We addressed it by manipulating lexical sharing directly by controlling which form-overlapping words received an embedding shared between languages versus language-specific embeddings.

We find that the bilingual LM keeps the two languages apart in the embedding space, but cross-lingual effects do occur via embedding sharing: friends (i.e.~cognate and loan words) show lower surprisal than language-unique control words in the conditions where either each form overlapping word is shared between Dutch and English, or only friends are shared between the two languages. Similarly, false friends have lower surprisal than controls either in conditions where sharing holds for all form overlapping words or just for false friends. These facilitation effects are largely explained by frequency-based signals.

The vocabulary condition that matches human data is one where only friends are shared between Dutch and English: the LM in this condition exhibits facilitation effects for friends without the corresponding facilitation for false friends. Other conditions fail to resemble human bilinguals with respect to processing the different word types: 1) the conditions with full overlap or just for false friends show facilitation for false friends, 2) the condition with minimal overlap (named entities and punctuation) does not show cross-lingual effects. Crucially, simulations of bilingual reading with LMs typically assume full surface overlap (see Section \ref{sec:rw_vocab}), which may bias conclusions about cross-lingual transfer and L2 learning.

Among the conditions tested, the one sharing only friends has the best \textit{descriptive adequacy} with respect to empirical findings of bilingual reading of overlapping words. However, it lacks \textit{explanatory adequacy} \citep{JacobsGrainger1994,Frank2021TowardCM}: the tokenizer and LM are based on explicitly coded rules about which forms are friends (and should each have a single embedding) and which are not. We cannot assume this is the mechanism behind cognate facilitation effect found in bilinguals. Importantly, humans learn meaning from grounded language learning, sensorimotor and social cues, and not (just) streams of text \citep{warstadt2022, chang-bergen-2022-word}. Yet, when the LM was left to train without explicit coding of (false) friends, it resulted in behavior that does not reflect that of bilingual humans. 

One of the open questions is why shared embeddings result in facilitation for false friends, rather than interference. One possibility is that surprisal in unconstrained sentence contexts mostly reflects the LM's ability to predict the surface form given the few semantic and lexical cues, and is as such mostly informed by frequency and distributional patterns rather than semantic competition between meanings. Another possibility might lie in the stimuli language (L1, so Dutch) and exposure. The LMs were largely trained on Dutch, while only a quarter of the data was in English. Perhaps the LM did not have a chance to properly learn the English meaning, and therefore does not exhibit interference effects when processing a sentence in Dutch. Generally, bilinguals tend to show cross-lingual effects in L2 reading and less so in L1 \citep{Titone2011BilingualLA}.

One of the main findings confirms the role of frequency in overlapping word forms specifically in the conditions with shared embeddings. When an embedding is shared, it is trained on data from both languages, which increases its effective frequency and typically reduces surprisal. This frequency benefit appears regardless of whether the word is a friend or a false friend. This suggests that overlap primarily provides a learning advantage through shared counts, not through semantic consistency.

The present study uses the GPT2-small Transformer architecture. Cross-lingual activation patterns in cognate facilitation have been observed across a range of architectures, including LSTM models \citep{winther-etal:2021-cumulative}, Transformers (albeit shallower than GPT2-small), and simple recurrent networks \citep{Roslund2022ModelingSP}. This suggests that cognate facilitation is driven less by architectural choice than by training data and procedure: specifically, unbalanced exposure (with greater L1 than L2 input) consistently gives rise to a cognate facilitation effect \citep{winther-etal:2021-cumulative, Roslund2022ModelingSP}. 

Furthermore, \citet{winther-etal:2021-cumulative} report that the order of L1 and L2 presentation plays a crucial role: the cognate effect emerges either when the model is first pretrained on L1 and subsequently trained on a mix of L1 and L2, or (in the absence of pretraining) when L1 data comprises the first 75\% of the epoch and L2 data the remainder. We follow the latter training procedure, but it would be valuable to test the former with different vocabulary designs as well. 

Our results can be situated within the framework of the BIA+ model \citep{Dijkstra2002TheAO}, which is one of the most influential computational accounts of bilingual word recognition. The model includes orthographic, phonological and semantic lexical representations with bidirectional connections between the representation types. In contrast, in our LMs, word representations are encoded at the embedding layer, where a word form either has a single (shared) or two language-specific embeddings. With this difference in mind, the vocabulary condition that best matches human bilingual data is broadly consistent with the BIA+ assumptions that cognates have a special shared representation due to their overlap in both form and meaning, while interlingual homographs (false friends) are represented by two separate orthographic entries. The role of form frequency in our results relates to the BIA+ model as well, where frequency is implemented as resting activation levels. Importantly, in BIA+, frequency and form overlap are separable sources of facilitation, while they are harder to disentangle in our LMs, where sharing an embedding both increases a word's effective frequency and enables cross-lingual transfer at the same time. Further comparisons between BIA+ and bilingual LMs, for instance using tasks that more directly probe semantic competition or that vary sentence context constraints, could shed more light on how well these two types of models align in explaining cross-lingual activation.

\section{Conclusion}
We presented a controlled study of vocabulary sharing in bilingual Dutch-English Transformer language models, targeting the question of whether bilingual LMs exhibit human-like cross-language activation. The results show that bilingual LMs generally keep Dutch and English contexts distinct, and cross-lingual effects arise primarily when the model shares embeddings. Under embedding sharing, overlapping forms become more predictable than language-unique controls, resulting in facilitation for both friends and false friends. This facilitation is largely explained by frequency and does not depend on whether the form-meaning mapping is shared across languages.

Among the evaluated vocabulary conditions, only the condition with shared embeddings for cognate and loan words 
reproduces the qualitative human pattern of facilitation for these words. Overall, these findings suggest that neural language models can reproduce some cross-lingual activation patterns, however, their alignment with bilingual reading behavior depends critically on how lexical overlap is treated in the encoding step.

\section{Limitations}

Our findings are currently based on a small-scale manipulation, which affected between 2.3\% and 4.3\% of the vocabulary depending on the condition. We did not explore how lexical interventions affect other aspects of cross-lingual transfer, e.g.~syntactic processing.

We tested the models on two sets of psycholinguistic stimuli, but only the \textit{friend study} \citep{Bultena2014CognateEI} has previously shown effects in bilinguals. The stimuli from the \textit{false-friend} experiment \citep{huisman-ma-thesis-ih} did not produce cross-lingual effects, limiting the extent to which model behavior can be compared to behavioral data. This points to a general challenge of scarcity of bilingual human responses to test the human-likeness of LMs. 

To measure cross-lingual effects in LMs, we use surprisal and similarity between representations, but other measures might capture patterns that correspond to human behavior, for example attention or the activation from the feed-forward networks \citep{oh-schuler-2022-entropy, kuribayashi-human-like-internally}. 

\section{Acknowledgements}
We thank Martin van Harmelen and Mayank Jobanputra for their help in implementing tokenizer and language model training. We also thank two anonymous reviewers for their constructive feedback.

Iza \v{S}krjanec is supported by the Konrad Zuse School of Excellence in Learning and Intelligent Systems (ELIZA) through the DAAD programme Konrad Zuse Schools of Excellence in Artificial Intelligence, sponsored by the Federal Ministry of Education and Research.

Iza \v{S}krjanec benefited from a Short-Term Scientific Mission funded by COST Action MultiplEYE (CA21131), supported by COST (European Cooperation in Science and Technology).

\section{Bibliographical References}\label{sec:reference}

\bibliographystyle{lrec2026-natbib}
\bibliography{lrec2026-example}



\section*{Appendix}
\setcounter{section}{0}
\renewcommand{\thesection}{\Alph{section}}
\section{Translations}

English translations of Dutch sentences in Section \ref{sec:exp_stimuli}.

\begin{itemize}
    \item [2a] The images of the fire ...
    \item [2b] The images of the bullet ...
\end{itemize}

\section{Language Model Hyperparameters}

The context window size was reduced to 256. We used gradient accumulation with  the effective batch size of 512. Weight decay was set to 0.1, the learning rate to 5e-4 with a cosine learning rate scheduler. The initial 1k steps of training were used as warm-up. All models were trained under the same random seed. 

\section{Vocabulary and LM Sizes} \label{app:vocab_lm_sizes}

\begin{table}[!h]
    \centering
    \begin{tabular}{|c|r|r|}
    \hline
        Condition & Vocabulary size & Num. parameters \\
        \hline
        A & 77,369 & 144,672,000 \\
        \hline
        B & 141,877 & 194,214,144\\
        \hline
        C & 144,170 & 195,975,168\\
        \hline
        D & 144,681 & 196,367,616 \\
        \hline
    \end{tabular}
    \caption{Vocabulary size and number of trainable parameters for the language model trained in each condition.}
    \label{tab:vocab_param_sizes}
\end{table}

\section{Training and Evaluation Loss} \label{app:loss_plots}
\begin{figure}[H]
    \centering
    \includegraphics[page=1,width=0.5\textwidth]{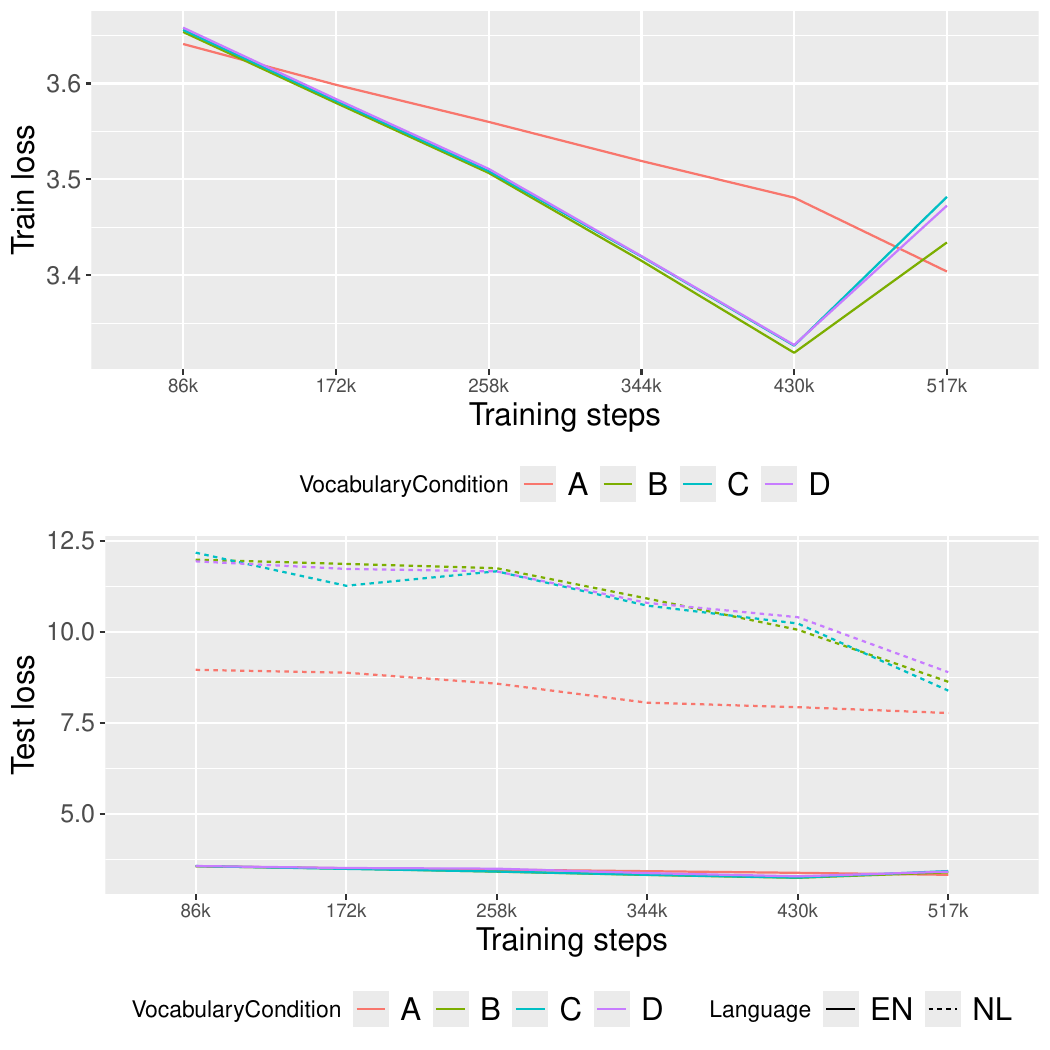}
    \caption{Training and evaluation loss.} 
    \label{fig:loss_curves}
\end{figure}

Figure \ref{fig:loss_curves} shows the loss values on the training and test sets at the end of each of 6 epochs in each condition. The training loss combines the values for the Dutch and English portion of the training set. We show separate values for the test set of each language. At the very end of each epoch, the LMs have been exposed to English more recently, so the test loss for English is quite low in all conditions, but high for Dutch, which indicates some level of attrition of Dutch. Compared to the other conditions, condition A (Full Overlap) shows a relatively higher training loss, but also lower test loss for Dutch. 

\section{ Diagrams for 5.1} \label{app:diagrams_vectors}

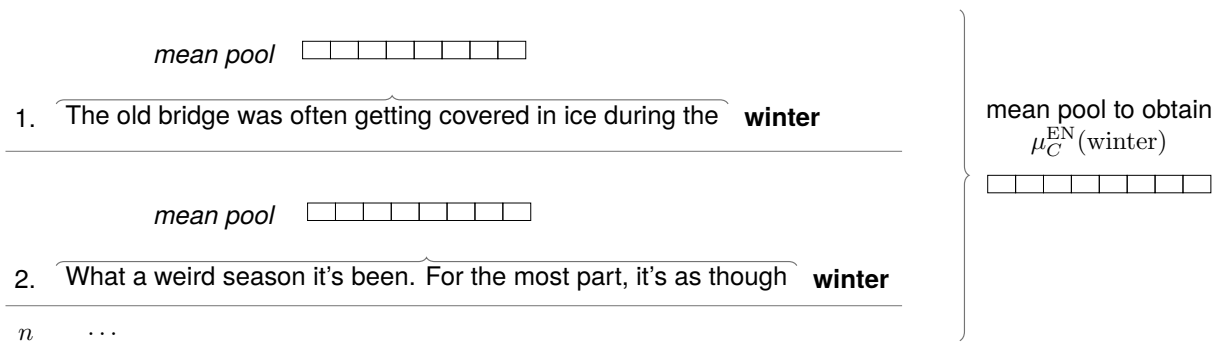
\begin{figure*}
\centering
\resizebox{\textwidth}{!}{%
\begin{tikzpicture}[
  box/.style={draw, minimum width=0.34cm, minimum height=0.22cm, inner sep=0pt, line width=0.35pt},
  line/.style={draw=black!55, line width=0.35pt},
  hbrace/.style={decorate, decoration={brace, amplitude=3.2pt}, draw=black!55, line width=0.45pt},
  vbrace/.style={decorate, decoration={brace, amplitude=3.6pt}, draw=black!55, line width=0.45pt}
]

\def\Hgap{1.45}
\def\TokN{8}
\def\dx{0.38}
\def\dy{0.24}
\pgfmathsetmacro{\Vw}{\TokN*\dx}

\def\CtxVecY{0.62}        
\def\CtxVecXshift{0.3}  
\def\PoolVecXshift{1.9}  
\def\PoolVecYshift{-0.9} 

\def\BX{13.0}

\node[anchor=west] at (1.9,1.55) {\itshape mean pool};

\node[anchor=west] (Aidx) at (0,0.72) {1.};
\node[anchor=west] (Astart) at ($(Aidx.east)+(0.15,0)$)
{The old bridge was often getting covered in ice during the};
\node[anchor=west] (Atok) at ($(Astart.east)+(0.10,0)$) {\textbf{winter}};

\draw[hbrace] ($(Astart.west)+(0,0.18)$) -- ($(Astart.east)+(0,0.18)$);
\coordinate (Acusp) at ($(Astart.west)!0.5!(Astart.east)+(0,0.18)$);

\coordinate (Acenter) at ($(Acusp)+(\CtxVecXshift,\CtxVecY)$);
\coordinate (Aleft)   at ($(Acenter)+(-\Vw/2,0)$);
\foreach \i in {0,...,7} {
  \draw[box] ($(Aleft)+(\i*\dx,0)$) rectangle ++(\dx,\dy);
}

\node[anchor=west] at (1.9,-\Hgap+0.80) {\itshape mean pool};

\node[anchor=west] (Bidx) at (0,-\Hgap-0.05) {2.};
\node[anchor=west] (Bstart) at ($(Bidx.east)+(0.15,0)$)
{What a weird season it's been. For the most part, it's as though};
\node[anchor=west] (Btok) at ($(Bstart.east)+(0.10,0)$) {\textbf{winter}};

\draw[hbrace] ($(Bstart.west)+(0,0.18)$) -- ($(Bstart.east)+(0,0.18)$);
\coordinate (Bcusp) at ($(Bstart.west)!0.5!(Bstart.east)+(0,0.18)$);

\coordinate (Bcenter) at ($(Bcusp)+(\CtxVecXshift-0.4,\CtxVecY)$);
\coordinate (Bleft)   at ($(Bcenter)+(-\Vw/2,0)$);
\foreach \i in {0,...,7} {
  \draw[box] ($(Bleft)+(\i*\dx,0)$) rectangle ++(\dx,\dy);
}

\coordinate (SepEnd) at ($(Btok.east)+(0.05,0)$);
\draw[line] (0,0.25) -- (SepEnd |- 0,0.25);

\coordinate (SepEndB) at ($(Btok.east)+(0.05,0)$);
\draw[line] (0,-\Hgap-0.42) -- (SepEndB |- 0,-\Hgap-0.42);


\node[anchor=west] (nmark) at ($(Bidx.west)+(0.05,-0.76)$) {$n$};
\node[anchor=west] (dots)  at ($(Bidx.east)+(0.45,-0.76)$) {$\cdots$};

\draw[vbrace] (\BX,2.19) -- (\BX,-\Hgap-0.9);

\coordinate (CUSP) at (\BX,0.58);

\coordinate (Pcenter) at ($(CUSP)+(\PoolVecXshift,\PoolVecYshift)$);
\coordinate (Pleft)   at ($(Pcenter)+(-\Vw/2,0)$);
\foreach \i in {0,...,7} {
  \draw[box] ($(Pleft)+(\i*\dx,0)$) rectangle ++(\dx,\dy);
}

\node[align=center] at ($(Pcenter)+(0,0.9)$)
{mean pool to obtain\\
$\mu^{\mathrm{EN}}_{C}(\mathrm{winter})$};

\end{tikzpicture}%
}
\caption{Mean pooling over the embeddings in the preceding tokens withing each sentence ($n=500$). Mean pooling over these to obtain a context embeddings from English samples $\mu_{C}^{EN}$ for the token \textit{winter}. The same was performed over Dutch sentences to obtain $\mu_{C}^{NL} (winter)$. }
\label{fig:meanpool-context}
\end{figure*}

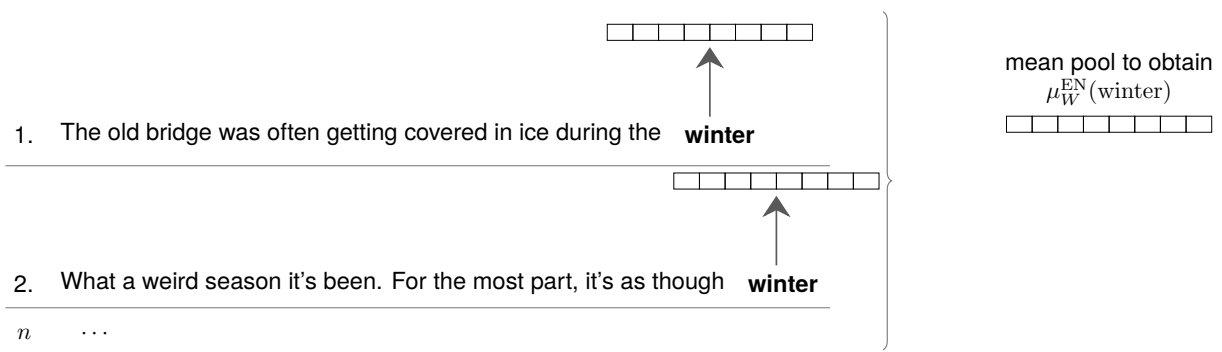
\begin{figure*}
\centering
\resizebox{\textwidth}{!}{%
\begin{tikzpicture}[
  box/.style={draw, minimum width=0.34cm, minimum height=0.22cm, inner sep=0pt, line width=0.35pt},
  line/.style={draw=black!55, line width=0.35pt},
  vbrace/.style={decorate, decoration={brace, amplitude=3.6pt}, draw=black!55, line width=0.45pt},
  upArrow/.style={-{Stealth[length=3.2mm,width=4.2mm]}, draw=black!65, line width=0.7pt}
]

\def\Hgap{1.45}
\def\TokN{8}
\def\dx{0.38}
\def\dy{0.24}
\pgfmathsetmacro{\Vw}{\TokN*\dx}

\def\VecY{1.15}        
\def\VecXshiftA{-0.15} 
\def\VecXshiftB{-0.10} 

\def\BX{13.0}          
\def\PoolX{3.35}       
\def\PoolY{0.75}       
\def\LabelY{1.55}      

\node[anchor=west] (Aidx) at (0,0.72) {1.};
\node[anchor=west] (Astart) at ($(Aidx.east)+(0.15,0)$)
{The old bridge was often getting covered in ice during the};
\node[anchor=west] (Atok) at ($(Astart.east)+(0.10,0)$) {\textbf{winter}};

\coordinate (Acenter) at ($(Atok.north)+(\VecXshiftA,\VecY)$);
\coordinate (Aleft)   at ($(Acenter)+(-\Vw/2,0)$);
\foreach \i in {0,...,7} {
  \draw[box] ($(Aleft)+(\i*\dx,0)$) rectangle ++(\dx,\dy);
}
\draw[upArrow] ($(Atok.north)+(\VecXshiftA,0.01)$) -- ($(Acenter)+(0,-0.1)$);

\node[anchor=west] (Bidx) at (0,-\Hgap-0.05) {2.};
\node[anchor=west] (Bstart) at ($(Bidx.east)+(0.15,0)$)
{What a weird season it's been. For the most part, it's as though};
\node[anchor=west] (Btok) at ($(Bstart.east)+(0.10,0)$) {\textbf{winter}};

\coordinate (Bcenter) at ($(Btok.north)+(\VecXshiftB,\VecY)$);
\coordinate (Bleft)   at ($(Bcenter)+(-\Vw/2,0)$);
\foreach \i in {0,...,7} {
  \draw[box] ($(Bleft)+(\i*\dx,0)$) rectangle ++(\dx,\dy);
}
\draw[upArrow] ($(Btok.north)+(\VecXshiftB,0.01)$) -- ($(Bcenter)+(0,-0.1)$);

\coordinate (SepEnd) at ($(Btok.east)+(0.05,0)$);
\draw[line] (0,0.25) -- (SepEnd |- 0,0.25);

\draw[line] (0,-\Hgap-0.42) -- (SepEnd |- 0,-\Hgap-0.42);

\node[anchor=west] (nmark) at ($(Bidx.west)+(0.05,-0.76)$) {$n$};
\node[anchor=west] (dots)  at ($(Bidx.east)+(0.45,-0.76)$) {$\cdots$};

\draw[vbrace] (\BX,2.55) -- (\BX,-\Hgap-1.05);

\coordinate (Pcenter) at (\BX+\PoolX,\PoolY);
\coordinate (Pleft)   at ($(Pcenter)+(-\Vw/2,0)$);
\foreach \i in {0,...,7} {
  \draw[box] ($(Pleft)+(\i*\dx,0)$) rectangle ++(\dx,\dy);
}

\node[align=center] at (\BX+\PoolX,\LabelY)
{mean pool to obtain\\
$\mu^{\mathrm{EN}}_{W}(\mathrm{winter})$};

\end{tikzpicture}%
}
\caption{Mean pooling over the token representations of the target word in $n$ English sentences ($n=500$) to obtain $\mu_{W}^{EN} (winter)$. Similarly, $n$ Dutch sentences are used to calculate $\mu_{W}^{NL} (winter)$.}
\label{fig:meanpool-token}
\end{figure*}

Figure \ref{fig:meanpool-context} illustrates mean pooling over the preceding context of target words to obtain $\mu_{C}$. Figure \ref{fig:meanpool-token} shows how we obtained the mean-pooled token representation $\mu_{W}$.



\end{document}